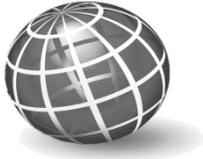

# The Journal of **Macro**Trends in Applied Science

MACROJOURNALS

# Improving Gravitational Search Algorithm Performance with Artificial Bee Colony Algorithm for Constrained Numerical Optimization

**Hasan Ali AKYÜREK\*, Ömer Kaan BAYKAN \*\*, Barış KOÇER\*\***
*\*Necmettin Erbakan University, School of Applied Sciences, Department of Management Information Sciences*
*\*\*Selcuk University, Faculty of Engineering, Department of Computer Engineering*

## Abstract

*In this paper, we propose an improved gravitational search algorithm named GSABC. The algorithm improves gravitational search algorithm (GSA) results improved by using artificial bee colony algorithm (ABC) to solve constrained numerical optimization problems. In GSA, solutions are attracted towards each other by applying gravitational forces, which depending on the masses assigned to the solutions, to each other. The heaviest mass will move slower than other masses and gravitate others. Due to nature of gravitation, GSA may pass global minimum if some solutions stuck to local minimum. ABC updates the positions of the best solutions that has obtained from GSA, preventing the GSA from sticking to the local minimum by its strong searching ability. The proposed algorithm improves the performance of GSA. The proposed method tested on 23 well-known unimodal, multimodal and fixed-point multimodal benchmark test functions. Experimental results show that GSABC outperforms or performs similarly to five state-of-the-art optimization approaches.*

I. INTRODUCTION

The meta-heuristic algorithms quickly determine the most suspicious parts of the solution space that may contain good results, but they are weaker in fine tuning of the best solution. To cope with the problem, meta-heuristic algorithms are used in combination with other methods to complete weak points of each other. It is claimed in [1] using a local optimization method on the results found by the meta-heuristic algorithm improves the success.

A hybrid algorithm is an algorithm that solves the same problem by combining two or more algorithms by choosing one from among the algorithms, or by switching between them during the flow. Beyond the combining "hybrid algorithm" can also be considered as compounding the best parts of many algorithms. Proposed method improves the results of the gravitational search algorithm (GSA) using the artificial bee colony algorithm(ABC).

II. MATERIAL AND METHODS

*A. Gravitational Search Algorithm*

Gravitational Search Algorithm(GSA) is an optimization algorithm based on Newton's universal gravitational law which proposed by Rashedi et al. 2009. The GSA, inspired by universal gravitational and motion laws, has an efficient computing capability [2,3].

In GSA, the performance of each solution is shown by the mass of the solution. Each mass gravitates the other masses in the search space with universal gravitational force. Thus, there is interactions between the masses. This gravitation allows all masses to move towards the heaviest mass. Therefore, the masses move together in the direction of gravitational forces.

The heaviest mass along the algorithm will move slower than other masses and gravitate others. When the termination criteria provided, the solution with the largest mass considered as the optimum solution.

In the GSA part of the algorithm, startup value of the gravity constant ($G$), number of candidate solutions and a certain search space are determined. The position of the $i$th candidate solution in the search space consisting of $N$ candidate solutions is expressed by equation (1).

$$x_i = (x_i^1, \ldots, x_i^d, \ldots, x_i^n) for, i = 1,2,\ldots,N \quad \#(1)$$

The value assigned to the universal gravitational constant determined at the beginning of the algorithm needs to be reduced iteration by iteration to control the search speed. The gravitational constant at $t$ is calculated by equation (2).

$$G(t) = G_0 * e^{-a\frac{t}{T}} \quad \#(2)$$

where $G_0$ represents the initial value of gravitational constant, $a$ is the constant value, $t$ is the number of iterations, and $T$ is the maximum number of iterations.

The best and worst solution values are obtained using equations (3) and (4) respectively.

$$best(t) = \min_{i=1 \to n} \left( \text{fit}_i(t) \right) \quad \#(3)$$

$$worst(t) = \max_{i=1 \to n} \left( \text{fit}_i(t) \right) \quad \#(4)$$

where $fit_i(t)$ is the fitness of $i$th mass, $best(t)$ is the best solution, and $worst(t)$ is the worst solution at $t$.

All the masses are calculated using Equations (5), (6) and (7), with the active gravitational mass ($M_{ai}$), passive gravitational mass ($M_{pi}$) and inertial mass ($M_{ii}$) being equal.

$$M_{ai} = M_{pi} = M_{ii} = M_i \#(5)$$

$$m_i = \frac{fit_i(t) - worst(t)}{best(t) - worst(t)} \#(6)$$

$$M_i(t) = \frac{m_i(t)}{\sum_{j=1}^{N} m_j(t)} \#(7)$$

Firstly, Euclidean distance between the two point masses is calculated by (8) then total force calculated by (9).

$$R_{ij}(t) = \|x_i(t), x_j(t)\|_2 \#(8)$$

The force between the masses is calculated using the distance between the masses.

$$F_{ij}^d(t) = G(t) \frac{M_{pi}(t) * M_{aj}(t)}{R_{ij}(t) + \varepsilon} \left(x_j^d(t) - x_i^d(t)\right) \#(9)$$

The accelerations of the masses are calculated by Equations (10) and forces between masses by (11).

$$a_i^d(t) = \frac{F_i^d(t)}{M_{ii}(t)} \#(10)$$

$$a_i(t) = G(t) \sum_{j=1}^{n} rand_j \frac{M_j(t)}{R_{ij}(t) + \varepsilon} (x_j - x_i) \#(11)$$

The masses give mutual acceleration as a result of the interaction. In Equation (12), the sum of the velocity at that moment and the velocity change at that moment is then calculated.

$$v_i^d(t+1) = rand_i * v_i^d(t) + a_i^d(t) \#(12)$$

The position of each mass in the system with the velocity change is updated with Equation (13).

$$x_i(t+1) = x_i(t) + v_i(t+1) \#(13)$$

Gravitational Search Algorithm Steps:
1. Initialize parameters and create random solutions.
2. Calculate fitness.
3. Update best and worst by using equation (3) and (4).
4. Calculate masses with equation (7).
5. Calculate gravity for solutions by using equation (9).
6. Calculate acceleration for solutions by using equation (11).
7. Update velocity and position of each solution with equation (12) and (13).
8. If the termination criteria provided, then the algorithm will stop otherwise the algorithm returns to Step 2, where the gravitational constant of the algorithm is updated and the search is continued until the termination criterion is provided.

Gravitational search algorithm flowchart is show in Figure 1.

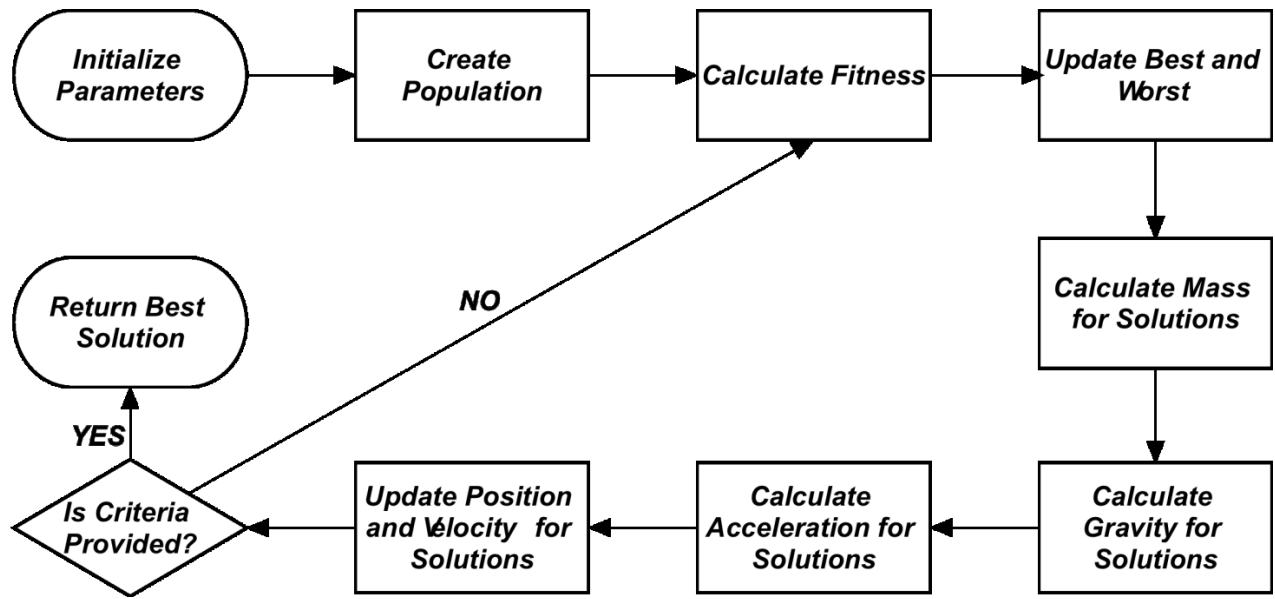

Figure 1. Gravitational Search Algorithm Flowchart

B. *Artificial Bee Colony*

The Artificial Bee Colony (ABC) Algorithm is one of the swarm intelligence based algorithms which proposed by Karaboga and Akay, 2009 [4,5,6]. The Artificial Bee Colony has the ability job sharing and self-organization as well as other swarm-based algorithms. This algorithm seeks the best solution through global and local search spaces according to the neighborhood principle [7]. The colony is divided into three groups:
1- Employed bees: Employed bees search for food sources that have more nectars based on the neighborhood principle. Each food source registered to an employed bee. Thus, the number of employed bees equals the number of food sources.
2- Onlooker bees: The onlooker bees wait in the nest and head towards the source according to amount of nutrient.
3- Scout Bees: If nutrient finishes in a food source, employed bee becomes scout bee start to search for food at random location.

Artificial Bee Colony Steps:
1- Set Initial Parameters: In this step, random sources of nutrients are generated in the environment and initial parameters are determined. Equation (14) is shown in the algorithm, which corresponds to generating a random initial value between the upper and lower bounds of the parameter.

$$x_{ij} = x_j^{min} + rand(0,1)\left(x_j^{max} - x_j^{min}\right) \#(14)$$

2- Calculation of Fitness Value: At this step, the fitness value of each foods is calculated.
3- Forward Employed Bees: At this step the worker bees are randomly distributed to food sources and dancing to identify new sources. The identification of new sources is based on neighborhood principle. Worker bee establishes a new source of food in the vicinity of the food source.

$$v_{ij} = x_{ij} + \emptyset_{ij}(x_{ij} - x_{kj}) \#(15)$$

A fitness value is assigned to the parameter vectors v representing a new source and a greedy selection operation is performed. $f_i$ is the error value produced by the $i$th solution. Equation (16) is used to calculate the fitness value of the solution. If the new food source is better than the old food source, the position old food source is deleted from the memory, and the position of new food source is stored.

$$fitness_i = \begin{cases} 1/(1+f_i), & f_i \geq 0 \\ 1 + abs(f_i), & f_i < 0 \end{cases} \#(16)$$

4- Forward Onlooker Bees: When the employed bees return to the nest, the onlooker bees communicate information about the quality of the food source. Accordingly, the onlooker bees choose their own food sources. Food sources with high nectar quality (fitness value) are more likely to be selected. The probability of selecting food sources is shown in Equation 17.

$$p_i = \frac{fitness_i}{\sum_{i=1}^{N} fitness_i} \#(17)$$

5- Termination Criteria Check: If the termination criteria is provided then the best solution is returned, else progress will continue.
6- Failure Control: If food sources can't be improved then algorithm progress step 7 otherwise algorithm will return to step 3.
7- Forward Scout Bees: If there is a food source that cannot be improved, it is necessary to release that food source and find a new food source instead. The scout bee is assigned to find this food source and randomly finds a new food source and transfers the information to the nest. Unless the termination criteria is provided, it will return to step 3.

Artificial bee colony algorithm flowchart is show in Figure 2.

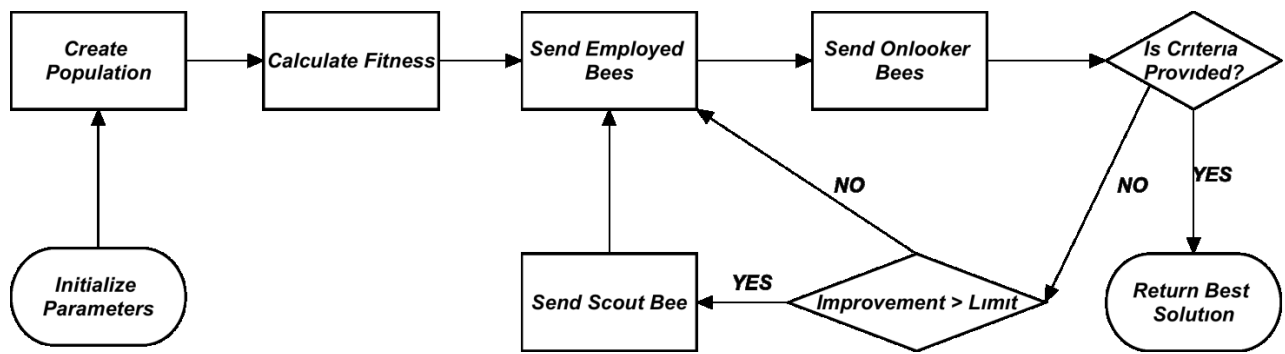

Figure 2. Artificial Bee Colony Algorithm Flowchart

III. HYBRID GSABC ALGORITHM

The proposed method is mainly based on the improvement of the results of the gravitational search algorithm (GSA) using the artificial bee colony algorithm(ABC). The half of the best population found by gravitational search algorithm sent to the artificial bee colony algorithm to start employed bees. Algorithm steps are given below.

1. Initialize parameters.
2. Create starting population.
3. Calculate fitness for current solution set.
4. Calculate Mass, Forces and Acceleration for each solution.
5. Update velocities and positions of candidate solutions.
6. Calculate fitness for updated solution set.
7. Select best number of populations/2 from solution set.
8. Send employed bees to food sources(solutions).
9. Send onlooker bees to food sources provided by employed bees.
10. If termination criteria provided return best solution else return step 4 and run until termination criteria provided.

Proposed method flowchart is show in Figure 3.

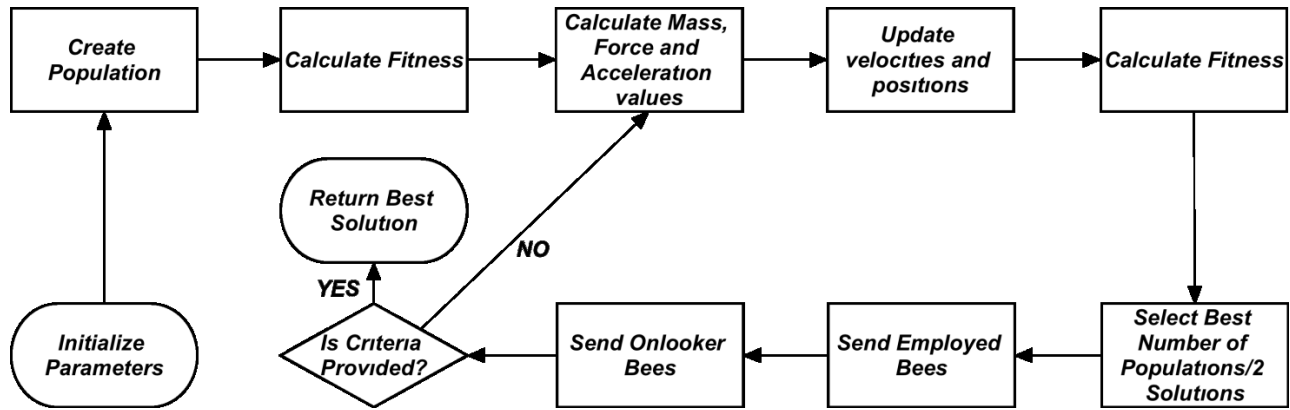

Figure 3. Proposed Method Flowchart

IV. EXPERIMENTAL RESULTS

Proposed method applied to 23 test functions [8-10] for Constrained Numerical Optimization. Proposed method compared with other optimization algorithms such Particle Swarm Optimization(PSO) [11], Differential Evolution(DE) [12,13], Fast Evolutionary Programing(FEP) [14], Each algorithm run 25 times and average values and standard deviations are reported in this section. Algorithms were run on the basis of the equal number of runs of fitness function.

Table 1. shows unimodal benchmark functions used in experiments.

| Unimodal Benchmark Functions | Dim | Range | $f_{min}$ |
|---|---|---|---|
| $f_1(x) = \sum_{i=1}^{n} x_i^2$ | 30 | [-100,100] | 0 |
| $f_2(x) = \sum_{i=1}^{n} |x_i| + \prod_{i=1}^{n} |x_i|$ | 30 | [-10,10] | 0 |
| $f_3(x) = \sum_{i=1}^{n} \left(\sum_{j=1}^{i} x_j\right)^2$ | 30 | [-100,100] | 0 |
| $f_4(x) = \max_i\{|x_i|, 1 \leq i \leq n\}$ | 30 | [-100,100] | 0 |
| $f_5(x) = \sum_{i=1}^{n-1}[100 * (x_{i+1} - x_i^2)^2 + (x_i - 1)^2]$ | 30 | [-30,30] | 0 |
| $f_6(x) = \sum_{i=1}^{n} ([x_i + 0.5])^2$ | 30 | [-100,100] | 0 |
| $f_7(x) = \sum_{i=1}^{n} i * x_i^4 + random[0,1)$ | 30 | [-1.28,1.28] | 0 |

Table 1. Unimodal Benchmark Functions

2-D Illustration of Unimodal benchmark functions shown in Figure 4.

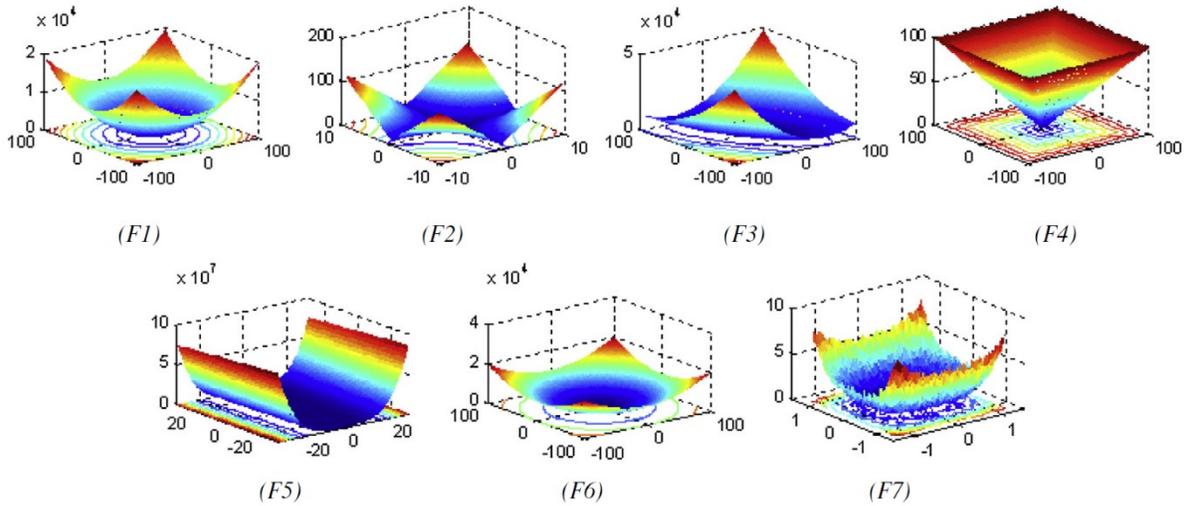

Figure 4. 2-D Illustration of Unimodal benchmark functions

Experimental results on unimodal benchmark functions shown in Table 2.

| $f$ | GSA | | ABC | | PSO | | DE | | FEP | | GSABC | |
|---|---|---|---|---|---|---|---|---|---|---|---|---|
| | Ave | Std | Ave | Std | Ave | Std | Ave | Std | Ave | Std | Ave | Std |
| $f_1$ | 9,855E-17 | 6,956E-17 | 5,388E-16 | 1,014E-16 | 1,360E-04 | 2,020E-04 | 8,200E-14 | 5,900E-14 | 5,700E-04 | 1,300E-04 | **5,008E-26** | 1,256E-26 |
| $f_2$ | 2,912E-08 | 8,024E-09 | 3,060E-10 | 1,305E-10 | 4,214E-02 | 4,542E-02 | 1,500E-09 | 9,900E-10 | 8,100E-03 | 7,700E-04 | **9,033E-13** | 7,827E-13 |
| $f_3$ | 9,155E+01 | 7,534E+01 | 4,996E+03 | 1,653E+03 | 7,013E+01 | 2,212E+01 | **6,800E-11** | 7,400E-11 | 1,600E-02 | 1,400E-02 | 2,993E-04 | 2,912E-04 |
| $f_4$ | 5,376E+00 | 1,326E+00 | 2,651E+01 | 6,718E+00 | 1,086E+00 | 3,170E-01 | **0,000E+00** | 0,000E+00 | 3,000E-01 | 5,000E-01 | 5,104E-09 | 1,039E-09 |
| $f_5$ | 2,671E+01 | 3,903E+00 | 5,126E-03 | 1,282E-01 | 9,672E+01 | 6,012E+01 | **0,000E+00** | 0,000E+00 | 5,060E+00 | 5,870E+00 | 2,686E-03 | 2,136E-02 |
| $f_6$ | 4,235E-17 | 5,050E-17 | 5,152E-16 | 1,194E-16 | 1,020E-04 | 8,280E-05 | **0,000E+00** | 0,000E+00 | **0,000E+00** | 0,000E+00 | 1,351E-25 | 8,538E-26 |
| $f_7$ | 1,055E-02 | 4,892E-03 | 1,328E-01 | 2,995E-02 | 1,229E-01 | 4,496E-02 | **4,630E-03** | 1,200E-03 | 1,415E-01 | 3,522E-01 | 5,956E-03 | 6,196E-03 |

Table 2. Experimental results on Unimodal Benchmark Functions

Table 3. shows multimodal benchmark functions used in experiments.

| Multimodal Benchmark Functions | Dim | Range | $f_{min}$ |
|---|---|---|---|
| $f_8(x) = \sum_{i=1}^{n} -x_i * \sin\left(\sqrt{|x_i|}\right)$ | 30 | [-500,500] | -12569.5 |
| $f_9(x) = \sum_{i=1}^{n} [x_i^2 - 10 * \cos(2\pi x_i) + 10]$ | 30 | [-5.12,5.12] | 0 |
| $f_{10}(x) = \sum_{i=1}^{n} -20 * \exp\left(-0.2\sqrt{\frac{1}{n}\sum_{i=1}^{n} x_i^2}\right) - \exp\left(\frac{1}{n}\sum_{i=1}^{n} cos(2\pi x_i)\right) + 20 + e$ | 30 | [-32,32] | 0 |
| $f_{11}(x) = \sum_{i=1}^{n} \frac{1}{4000}\sum_{i=1}^{n} x_i^2 - \prod_{i=1}^{n} \cos\left(\frac{x_i}{\sqrt{i}}\right) + 1$ | 30 | [-600,600] | 0 |
| $f_{12}(x) = \sum_{i=1}^{n} \frac{\pi}{n}\Big\{10\sin(\pi y_i) + \sum_{i=1}^{n-1}(y_i - 1)^2[1 + 10\sin^2(\pi y_{i+1})] + (y_n - 1)^2\Big\}$ $+ \sum_{i=1}^{n} u(x_i, 10, 100, 4)$ $y_i = 1 + \frac{x_i + 1}{4}$ $u(x_i, a, k, m) = \begin{cases} k(x_i - a)^m & x_i > a \\ 0 & -a < x_i < a \\ k(-x_i - a)^m & x_i < -a \end{cases}$ | 30 | [-50,50] | 0 |
| $f_{13}(x) = 0.1\Big\{\sin(3\pi x_1) + \sum_{i=1}^{n}(x_i - 1)^2[1 + \sin^2(3\pi x_i + 1)]$ $+ (x_n - 1)^2[1 + \sin^2(2\pi x_n)]\Big\} + \sum_{i=1}^{n} u(x_i, 5, 100, 4)$ | 30 | [-50,50] | 0 |

Table 3. Multimodal Benchmark Functions

2-D Illustration of Multimodal benchmark functions shown in Figure 5.

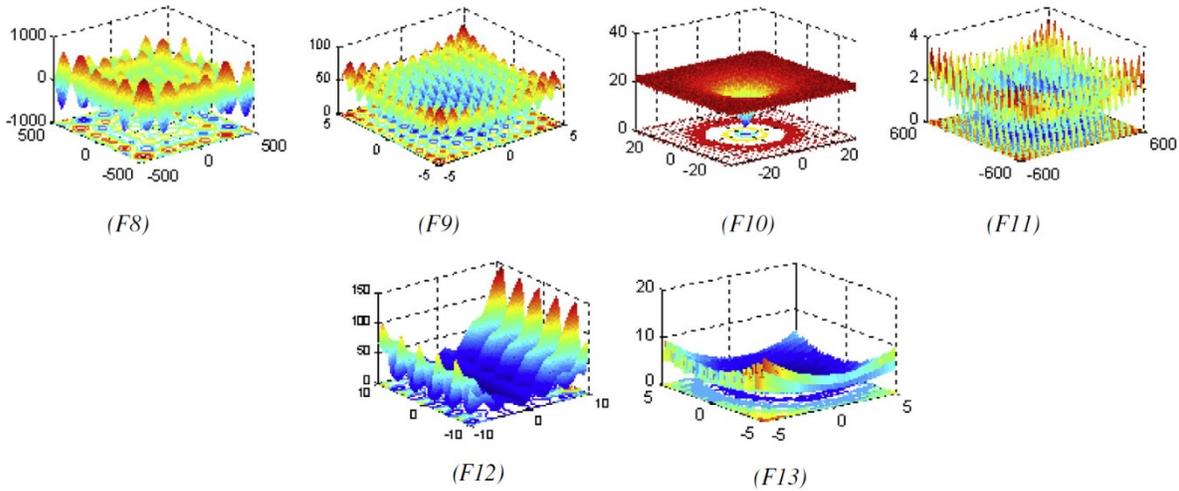

Figure 5. 2-D Illustration of Multimodal benchmark functions

Experimental results on unimodal benchmark functions shown in Table 4.

| $f$ | GSA | | ABC | | PSO | | DE | | FEP | | GSABC | |
|---|---|---|---|---|---|---|---|---|---|---|---|---|
| | Ave | Std | Ave | Std | Ave | Std | Ave | Std | Ave | Std | Ave | Std |
| $f_8$ | -3,719E+03 | 5,573E+02 | **-1,257E+04** | 4,183E+01 | -4,841E+03 | 1,153E+03 | -1,108E+04 | 5,747E+02 | -1,255E+04 | 5,260E+01 | -1,221E+04 | 1,266E+02 |
| $f_9$ | 1,198E+01 | 2,229E+00 | 7,215E+00 | 4,168E+00 | 4,670E+01 | 1,163E+01 | 6,920E+01 | 3,880E+01 | 4,600E-02 | 1,200E-02 | **0,000E+00** | 1,004E-11 |
| $f_{10}$ | 6,825E-09 | 1,131E-09 | 5,467E-09 | 2,154E-09 | 2,760E-01 | 5,090E-01 | 9,700E-08 | 4,200E-08 | 1,800E-02 | 2,100E-03 | **1,927E-13** | 2,554E-13 |
| $f_{11}$ | 1,444E+00 | 6,575E-01 | **0,000E+00** | 2,294E-12 | 9,215E-03 | 7,724E-03 | **0,000E+00** | 0,000E+00 | 1,600E-02 | 2,200E-02 | **0,000E+00** | 6,651E-01 |
| $f_{12}$ | 6,853E-17 | 3,051E-02 | 5,014E-16 | 1,105E-16 | 6,917E-03 | 2,630E-02 | 7,900E-15 | 8,000E-15 | 9,200E-06 | 3,600E-06 | **6,954E-28** | 8,446E-13 |
| $f_{13}$ | 7,172E-16 | 2,205E-04 | 4,620E-16 | 1,353E-16 | 6,675E-03 | 8,907E-03 | 5,100E-14 | 4,800E-14 | 1,600E-04 | 7,300E-05 | **1,325E-26** | 5,621E-16 |

Table 4. Experimental results on Multimodal Benchmark Functions

Table 5. shows multimodal benchmark functions used in experiments.

| Fixed-dimension Multimodal Benchmark Functions | Dim | Range | $f_{min}$ |
|---|---|---|---|
| $f_{14}(x) = \left( \frac{1}{500} + \sum_{j=1}^{25} \frac{1}{j + \sum_{i=1}^{2}(x_i - a_{ij})^6} \right)^{-1}$ | 2 | [-65,65] | 1 |
| $f_{15}(x) = \sum_{i=1}^{11} \left[ a_i - \frac{x_1(b_i^2 + b_i x_2)}{b_i^2 + b_i x_3 + x_4} \right]^2$ | 4 | [-5,5] | 0.00030 |
| $f_{16}(x) = 4x_1^2 - 2.1x_1^4 + \frac{1}{3}x_1^6 + x_1 x_2 - 4x_2^2 + 4x_2^4$ | 2 | [-5,5] | -1.0316 |
| $f_{17}(x) = \left( x_2 - \frac{5.1}{4\pi^2}x_1^2 + \frac{5}{\pi}x_1 - 6 \right)^2 + 10\left(1 - \frac{1}{8\pi}\right)\cos x_1 + 10$ | 2 | [-5,5] | 0.398 |
| $f_{18}(x) = [1 + (x_1 + x_2 + 1)^2(19 - 14x_1 + 3x_1^2 - 14x_2 + 6x_1 x_2 + 3x_2^2)]x[30 + (2x_1 - 3x_2)^2 x(18 - 32x_1 + 12x_1^2 + 48x_2 - 36x_1 x_2 + 27x_2^2]$ | 2 | [-2,2] | 3 |
| $f_{19}(x) = -\sum_{i=1}^{4} c_i \exp\left( -\sum_{j=1}^{3} a_{ij}(x_j - p_{ij})^2 \right)$ | 3 | [1,3] | -3.86 |
| $f_{20}(x) = -\sum_{i=1}^{n} c_i \exp\left( -\sum_{j=1}^{6} a_{ij}(x_j - p_{ij})^2 \right)$ | 6 | [0,1] | -3.32 |
| $f_{21}(x) = -\sum_{i=1}^{5} [(X - a_i)(X - a_i)^T + c_i]^{-1}$ | 4 | [0,10] | -10.1532 |
| $f_{22}(x) = -\sum_{i=1}^{7} [(X - a_i)(X - a_i)^T + c_i]^{-1}$ | 4 | [0,10] | -10.4028 |
| $f_{23}(x) = -\sum_{i=1}^{10} [(X - a_i)(X - a_i)^T + c_i]^{-1}$ | 4 | [0,10] | -10.5363 |

Table 5. Fixed-Dimension Multimodal Benchmark Functions

2-D Illustration of Fixed-dimension multimodal benchmark functions shown in Figure 6.

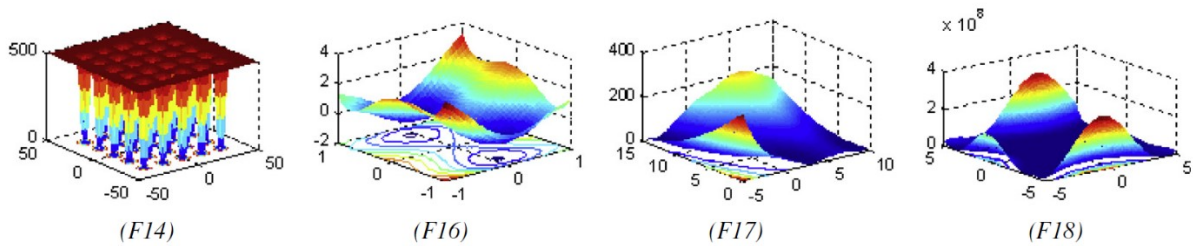

(F14)   (F16)   (F17)   (F18)

Figure 6. 2-D Illustration of Fixed-dimension multimodal benchmark functions

Experimental results on unimodal benchmark functions shown in Table 6.

| $f$ | GSA | | ABC | | PSO | | DE | | FEP | | GSABC | |
|---|---|---|---|---|---|---|---|---|---|---|---|---|
| | Ave | Std | Ave | Std | Ave | Std | Ave | Std | Ave | Std | Ave | Std |
| $f_{14}$ | 1,072E+00 | 2,971E+00 | **9,980E-01** | 0,000E+00 | 3,627E+00 | 2,561E+00 | 9,980E-01 | 3,300E-16 | 1,220E+00 | 5,600E-01 | **9,980E-01** | 1,364E+00 |
| $f_{15}$ | 1,370E-03 | 1,486E-03 | 6,613E-04 | 6,718E-04 | 5,770E-04 | 2,220E-04 | **4,500E-14** | 3,300E-04 | 5,000E-04 | 3,200E-04 | 3,333E-04 | 1,196E-04 |
| $f_{16}$ | -1,032E+00 | 3,864E-09 | -1,032E+00 | 0,000E+00 | **-1,032E+00** | 6,250E-16 | **-1,032E+00** | 3,100E-13 | -1,030E+00 | 4,900E-07 | -1,032E+00 | 0,000E+00 |
| $f_{17}$ | 3,979E-01 | 9,367E-08 | 3,979E-01 | 0,000E+00 | **3,979E-01** | 0,000E+00 | **3,979E-01** | 9,900E-09 | 3,980E-01 | 1,500E-07 | 3,979E-01 | 0,000E+00 |
| $f_{18}$ | 3,000E+00 | 8,529E-07 | **3,000E+00** | 4,575E-09 | 3,000E+00 | 1,330E-15 | 3,000E+00 | 2,000E-15 | 3,020E+00 | 1,100E-01 | **3,000E+00** | 2,735E-12 |
| $f_{19}$ | -3,863E+00 | 6,050E-03 | **-3,863E+00** | 4,885E-15 | -3,863E+00 | 2,580E-15 | N/A | N/A | -3,860E+00 | 1,400E-05 | **-3,863E+00** | 0,000E+00 |
| $f_{20}$ | -3,322E+00 | 3,299E-01 | **-3,322E+00** | 0,000E+00 | -3,266E+00 | 6,052E-02 | N/A | N/A | -3,270E+00 | 5,900E-02 | **-3,322E+00** | 2,851E-02 |
| $f_{21}$ | -1,015E+01 | 1,365E+00 | -1,015E+01 | 0,000E+00 | -6,865E+00 | 3,020E+00 | **-1,015E+01** | 2,500E-06 | -5,520E+00 | 1,590E+00 | -1,015E+01 | 1,661E+00 |
| $f_{22}$ | -1,040E+01 | 1,412E-01 | **-1,040E+01** | 0,000E+00 | -8,457E+00 | 3,087E+00 | -1,040E+01 | 3,900E-07 | -5,530E+00 | 2,120E+00 | **-1,040E+01** | 6,301E-01 |
| $f_{23}$ | -1,054E+01 | 5,302E-03 | **-1,054E+01** | 0,000E+00 | -9,953E+00 | 1,783E+00 | -1,054E+01 | 1,900E-07 | -6,570E+00 | 3,140E+00 | **-1,054E+01** | 0,000E+00 |

Table 6. Experimental results on Fixed-Dimension Multimodal Benchmark Functions

V. CONCLUSION AND FEATURE WORK

In this study, we propose an improved Gravitational Search Algorithm with Artificial Bee Colony Algorithm. The proposed algorithm improves the algorithm's performance. We test the presented method on 23 well-known benchmark test functions. Comparisons show that GSABC outperforms or performs similarly to five state-of-the-art approaches in terms of the quality of the resulting solutions. Our future work will apply the proposed method to engineering optimization problems.

REFERENCES

[1] Pei, S., Ouyang, A., & Tong, L. (2015). A hybrid algorithm based on bat-inspired algorithm and differential evolution for constrained optimization problems. *International Journal of Pattern Recognition and Artificial Intelligence*, *29*(04), 1559007.

[2] Rashedi, E., Nezamabadi-Pour, H., & Saryazdi, S. (2009). GSA: a gravitational search algorithm. *Information sciences*, *179*(13), 2232-2248.

[3] Rashedi, E., Nezamabadi-Pour, H., & Saryazdi, S. (2010). BGSA: binary gravitational search algorithm. *Natural Computing*, *9*(3), 727-745.

[4] Bonabeau, E., Dorigo, M., & Theraulaz, G. (1999). *Swarm intelligence: from natural to artificial systems* (No. 1). Oxford university press.

[5] Karaboga, D., & Akay, B. (2009). A comparative study of artificial bee colony algorithm. *Applied mathematics and computation*, *214*(1), 108-132.

[6] Karaboga, D., & Basturk, B. (2008). On the performance of artificial bee colony (ABC) algorithm. *Applied soft computing*, *8*(1), 687-697.

[7] Guo, P., Cheng, W., & Liang, J. (2011, July). Global artificial bee colony search algorithm for numerical function optimization. In *Natural Computation (ICNC), 2011 Seventh International Conference on* (Vol. 3, pp. 1280-1283). IEEE.

[8] Digalakis, J. G., & Margaritis, K. G. (2001). On benchmarking functions for genetic algorithms. *International journal of computer mathematics*, *77*(4), 481-506.

[9] Molga, M., & Smutnicki, C. (2005). Test functions for optimization needs. *Test functions for optimization needs*.

[10] Yang, X. S. (2010). Appendix A: test problems in optimization. *Engineering optimization*, 261-266.

[11] Kennedy, J., & Eberhart, R. (1942). Particle swarm optimization 1995 IEEE International Conference on Neural Networks Proceedings.

[12] Huang, F. Z., Wang, L., & He, Q. (2007). An effective co-evolutionary differential evolution for constrained optimization. *Applied Mathematics and computation*, *186*(1), 340-356.

[13] Storn, R., & Price, K. (1997). Differential evolution–a simple and efficient heuristic for global optimization over continuous spaces. *Journal of global optimization*, *11*(4), 341-359.

[14] Yao, X., Liu, Y., & Lin, G. (1999). Evolutionary programming made faster. *IEEE Transactions on Evolutionary computation*, *3*(2), 82-102.